\documentclass{article}

\usepackage{microtype}
\usepackage{graphicx}
\usepackage{subcaption}
\usepackage{booktabs} 
\usepackage{tabularx}
\usepackage{multirow}
\usepackage{array}

\usepackage{hyperref}

\usepackage[preprint]{icml2026}

\usepackage{amsmath}
\usepackage{amssymb}
\usepackage{mathtools}
\usepackage{amsthm}
\usepackage{bbm}

\usepackage[capitalize,noabbrev]{cleveref}

\theoremstyle{plain}

\theoremstyle{definition}

\theoremstyle{remark}

\usepackage{float}    
\usepackage{placeins} 

\icmltitlerunning{Domain Restriction via SAE Layer Transitions}
\begin{document}

\twocolumn[
\icmltitle{Domain Restriction via Multi SAE Layer Transitions}

\begin{icmlauthorlist}
  \icmlauthor{Elias Shaheen}{technion}
  \icmlauthor{Avi Mendelson}{technion}
\end{icmlauthorlist}

\icmlaffiliation{technion}{Technion -- Israel Institute of Technology, Haifa, Israel}
\icmlcorrespondingauthor{Elias Shaheen}{eliasshaheen@campus.technion.ac.il}
\icmlcorrespondingauthor{Avi Mendelson}{mendlson@technion.ac.il}

\icmlkeywords{Machine Learning, ICML}

\vskip 0.3in
]



\printAffiliationsAndNotice{}  

\begin{abstract}
  The general-purpose nature of Large Language Models (LLMs) presents a significant challenge for domain-specific applications, often leading to out-of-domain (OOD) interactions that undermine the provider's intent. Existing methods for detecting such scenarios treat the LLM as an uninterpretable black box and overlook the internal processing of inputs. In this work we show that layer transitions provide a promising avenue for extracting domain-specific signature. Specifically, we present several lightweight ways of learning on internal dynamics encoded using a sparse autoencoder (SAE) that exhibit great capability in distinguishing OOD texts. Building on top of SAEs representation transitions enables us to better interpret the LLM internal evolution of input processing and shed light on its decisions. We provide a comprehensive analysis of the method and benchmark it with the gemma-2 2B and 9B models. Our results emphasize the efficacy of the internal process in capturing fine-grained input-related details.\footnote{Code will be released in the future.}
\end{abstract}

\section{Introduction}
Large language models are increasingly used as interfaces for information access and automation in real-world systems \cite{chen2023frugalgptuselargelanguage,hu2024routerbenchbenchmarkmultillmrouting}. In many deployments, LLMs are embedded in agentic workflows that plan, call tools, and coordinate specialized components to accomplish user tasks.\cite{yao2023reactsynergizingreasoningacting, schick2023toolformerlanguagemodelsteach} To improve reliability and cost-efficiency, service providers often restrict these systems to a specific domain and deploy domain-specialized agents rather than a single general-purpose assistant\cite{chen2023frugalgptuselargelanguage,hu2024routerbenchbenchmarkmultillmrouting}. As a result, ensuring that requests are handled by the appropriate agent has become a key requirement for practical LLM-based systems.\cite{hu2024routerbenchbenchmarkmultillmrouting}

However, these systems introduce additional complexity: multiple components must coordinate through language and intermediate decisions.\cite{liu2025agentbenchevaluatingllmsagents, zhou2024webarenarealisticwebenvironment, yao2023reactsynergizingreasoningacting} Since LLM outputs can be unreliable and sometimes hallucinated, errors can cascade across steps and cause the workflow to diverge from the intended goal.\cite{Huang_2025, liu2024trustworthyllmssurveyguideline} This results in failed task execution, poor user experience, and wasted compute and operational overhead for deployers.

This motivates a gating mechanism that assesses whether an input request is appropriate for a given agent\cite{hu2024routerbenchbenchmarkmultillmrouting,chen2023frugalgptuselargelanguage, shazeer2017outrageouslylargeneuralnetworks}. Formally, given a request \(x\) and an agent scope \(S\), the goal is to decide whether \(x\) is in-scope (to be handled by the agent) or out-of-scope (to be rejected or rerouted). We view this as an out-of-distribution (OOD) detection problem in language, where in-scope requests follow the target distribution induced by \(S\) . Performance is evaluated using standard OOD discrimination metrics.\cite{hendrycks2018baselinedetectingmisclassifiedoutofdistribution, liang2020enhancingreliabilityoutofdistributionimage}

Prior work has studied OOD detection for neural networks and language models by leveraging different signals available in transformer architectures.\cite{podolskiy2022revisitingmahalanobisdistancetransformerbased, sun2022outofdistributiondetectiondeepnearest} 
A common post-hoc baseline uses output confidence, e.g., maximum softmax probability (MSP) \cite{hendrycks2018baselinedetectingmisclassifiedoutofdistribution}. 
Beyond output scores, many approaches exploit internal representations, using feature-space distances or fitted density models to separate in-distribution inputs from OOD samples \cite{lee2018simpleunifiedframeworkdetecting,podolskiy2022revisitingmahalanobisdistancetransformerbased,sun2022outofdistributiondetectiondeepnearest}. 
In the NLP setting, pretrained transformers have also been shown to substantially improve robustness to distribution shift and out-of-domain inputs \cite{hendrycks2020pretrainedtransformersimproveoutofdistribution,uppaal2023finetuningneededpretrainedlanguage}. 
For broader context and additional families of methods, we refer to recent surveys \cite{lu2025outofdistributiondetectiontaskorientedsurvey,lang2023surveyoutofdistributiondetectionnlp}.

While these methods can be effective, they often offer limited transparency into \emph{why} an input is judged as out-of-scope \cite{räuker2023transparentaisurveyinterpreting, elhage2022toymodelssuperposition}, and practical constraints may limit access to labeled OOD data or the ability to fine-tune models.\cite{lu2025outofdistributiondetectiontaskorientedsurvey} This motivates approaches that are both interpretable and lightweight to deploy, while maintaining strong OOD discrimination in realistic settings.

In this work, we investigate whether transformer internals provide a reliable signal for deciding in-scope versus out-of-scope requests. Concretely, we analyze the evolution of residual-stream representations and map them into an interpretable feature space using sparse autoencoders (SAEs).\cite{cunningham2023sparseautoencodershighlyinterpretable} Using only in-scope examples, we characterize a reference signature of the model’s internal behavior and use deviations from this signature to score inputs at test time. This design is lightweight and does not require labeled OOD data or fine-tuning of the base model. We evaluate the approach under both near- and far-OOD settings and provide analysis linking detection performance to internal representation dynamics.

Our contributions are:
\begin{itemize}
    \item We propose an \emph{ID-only} scope-gating method that detects out-of-scope text by modeling \emph{depthwise transitions} of sparse, interpretable SAE features, without fine-tuning the base LLM or using labeled OOD data.
    \item We introduce an SAE$\rightarrow$SDR representation for transformer-layer trajectories and a modular sequential scoring framework, and we compare lightweight backends (first-order Markov, Hierarchical Temporal Memory (HTM), and an RNN predictor) for anomaly scoring.
    \item We provide representation analyses on Gemma2-2B characterizing where domain-consistent structure emerges across depth and how transition-based modeling differs from static overlap.
    \item We evaluate the resulting gate on both near- and far-OOD benchmarks, and we provide interpretability analyses (including a hard-OOD boundary case study) linking detection behavior to decoded SAE feature transitions.
\end{itemize}


\section{Related Work and Background}
\subsection{Sparse Autoencoders}
\label{sec:sae_background}

\textbf{Motivation: sparse feature dictionaries for superposed representations.}
Transformer residual streams are high-dimensional and distributed: many distinct latent factors can be represented in overlapping directions, a phenomenon often described as \emph{superposition}. This makes single-neuron interpretations unreliable and motivates learning an explicit \emph{feature dictionary} that decomposes activations into a sparse set of (approximately) reusable components \cite{elhage2022toymodelssuperposition}.

\textbf{SAE formulation.}
Given a layer $\ell$ and a residual-stream activation $h_{\ell,t}\in\mathbb{R}^d$ at token position $t$, a sparse autoencoder (SAE) learns an encoder--decoder pair $(\mathrm{Enc}_\ell,\mathrm{Dec}_\ell)$ such that
\begin{equation}
z_{\ell,t}=\mathrm{Enc}_\ell(h_{\ell,t})\in\mathbb{R}^D,\qquad
\hat h_{\ell,t}=\mathrm{Dec}_\ell(z_{\ell,t}),
\end{equation}
with $D\gg d$ and $z_{\ell,t}$ encouraged to be sparse. Training typically minimizes a reconstruction loss plus a sparsity penalty, e.g.,
\begin{equation}
\min_{\mathrm{Enc}_\ell,\mathrm{Dec}_\ell}\;
\mathbb{E}\big[\|h_{\ell,t}-\hat h_{\ell,t}\|_2^2\big]
\;+\;\lambda\|z_{\ell,t}\|_1,
\end{equation}
or related sparsity constraints. Intuitively, $\mathrm{Dec}_\ell$ learns a set of dictionary atoms (columns) and the sparse code $z_{\ell,t}$ selects a small subset of them to explain $h_{\ell,t}$.

\textbf{Interpretability and empirical evidence.}
Recent work shows that SAEs trained on LLM activations can yield \emph{highly interpretable} features, often aligning with human-recognizable concepts, syntax, or task-related patterns, and improving over direct neuron-level inspection \cite{cunningham2023sparseautoencodershighlyinterpretable, paulo2025automaticallyinterpretingmillionsfeatures}. This line of work operationalizes the “features not neurons” view of mechanistic interpretability by replacing polysemantic neuron activations with sparse, compositional feature activations.

\textbf{Why SAEs are useful for our setting.}
Our goal is to detect \emph{domain restriction} signals from the \emph{internal dynamics} of an LLM using only in-domain data. SAEs provide two properties that are particularly useful here:
(i) a \emph{common coordinate system} across examples (feature indices) that makes trajectories comparable across inputs and layers \cite{marks2025sparsefeaturecircuitsdiscovering, Arad_2025}; and
(ii) a \emph{sparsified} representation where Top-$k$ active features can be treated as a discrete state suitable for transition modeling.\cite{Cui_2016}
This representation is suitable to lightweight sequential models (e.g., Markov/HTM/RNN backends) than dense residual vectors, and it supports post-hoc analysis by mapping active indices back to feature descriptions.

\textbf{Related directions and tooling.}
Several closely related efforts learn and analyze sparse feature dictionaries in transformers, including variations on dictionary learning and cross-layer feature mappings (e.g., \emph{crosscoders}) \cite{lindsey2024crosscoders}. On the tooling side, open-source libraries such as \texttt{SAELens} facilitate training and applying SAEs to modern LLMs and standardizing evaluation/visualization workflows \cite{bloom2024saetrainingcodebase}, additionally we use Neuronpedia for leveraging features data such as text label and global densities. \cite{neuronpedia}
In our work, we treat SAEs as a fixed, pretrained decomposition of each layer’s residual stream and focus on what can be learned \emph{on top} of these features from in-domain data alone---namely, whether \emph{depthwise transition regularities} form a stable, domain-specific signature suitable for scope gating.

\subsection{OOD Detection}
Deployed models inevitably get exposed to out-of-distribution samples, as opposed to the closed-world assumption \cite{NIPS1991_ff4d5fbb}, such scenario might cause model's to behave in ways that are not expected and even sometimes harmful, this motivates the detection of such outliers. Traditionally OOD detection was the paradigm of detecting test-time samples that were not present in the training data of the model \cite{sun2022outofdistributiondetectiondeepnearest,lee2018simpleunifiedframeworkdetecting,hendrycks2018baselinedetectingmisclassifiedoutofdistribution}, however and since LLMs are trained on extremely broad and heterogeneous web-scale corpora, the boundary between ID and OOD became blurry.\cite{lang2023surveyoutofdistributiondetectionnlp}
The feild of OOD detection splits into different settings based on the presence of OOD samples and labels for the training samples \cite{lang2023surveyoutofdistributiondetectionnlp}, the method this paper discusses operates in the absence of both. Naturally the field migrated into exploring detecting OOD in the inputs to Large pre-trained encoders like BERT\cite{hendrycks2020pretrainedtransformersimproveoutofdistribution, podolskiy2022revisitingmahalanobisdistancetransformerbased, chen2022holisticsentenceembeddingsbetter, devlin2019bertpretrainingdeepbidirectional, uppaal2023finetuningneededpretrainedlanguage}, we explore a similar paradigm in their counterparts, the decoder-only Large Pretrained Models.

Closest to our experimental setting is \cite{zhang2025finetunedlargelanguagemodel}, with the difference of them using a fine-tune variant of the model to calculate the anomaly score, whereas we adapt an auxilary lightweight model on the internal activation transitions to do so.

\section{Methodology}
\label{sec:methodology}

\begin{figure*}[t]
  \centering
  \includegraphics[
    width=0.9\textwidth,
    trim=1cm 15.1cm 0cm 2.4cm, 
    clip
  ]{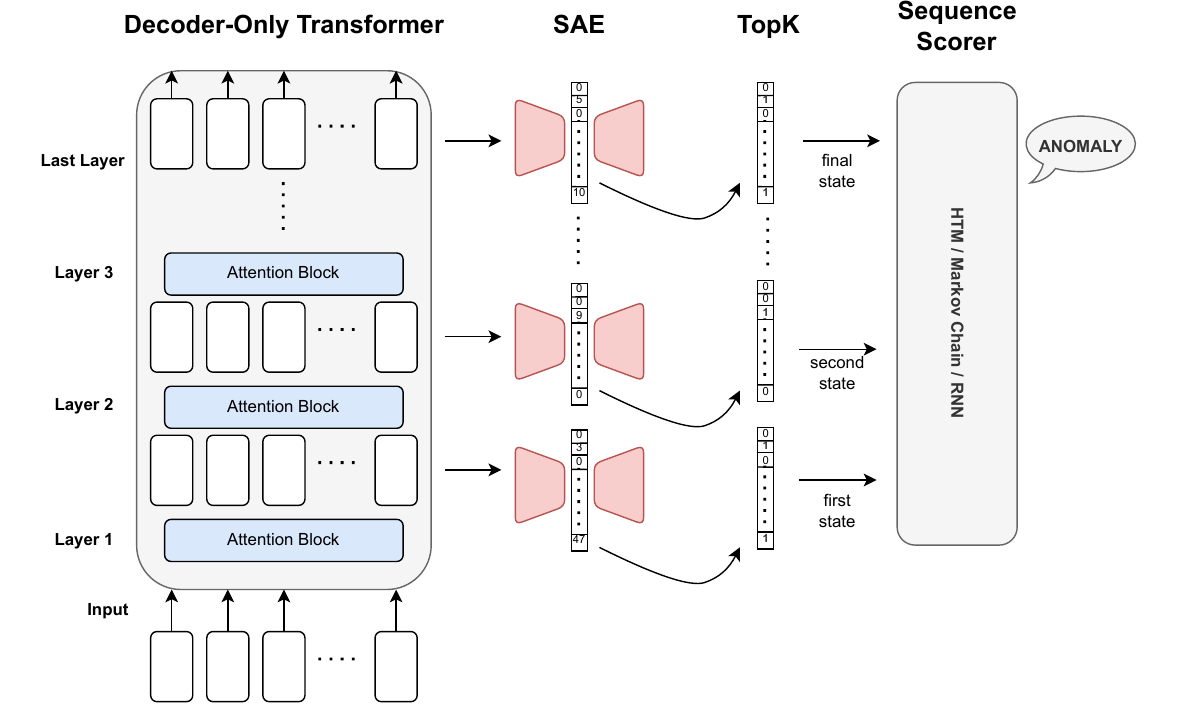}
  \caption{
Pipeline overview. Given an input, we extract residual-stream activations across layers, encode them with layer-wise SAEs, pool over tokens, mask globally frequent features, and binarize via Top-$k$ to obtain an SAE$\rightarrow$SDR depth trajectory.
A lightweight sequential scorer (Markov/HTM/RNN) trained on ID-only data assigns an anomaly score from depthwise transitions.
}

  \label{fig:diagram1}
\end{figure*}

Our method, illustrated in Figure~\ref{fig:diagram1}, is a four-stage pipeline designed to detect out-of-domain inputs by analyzing the internal activation flow of an LLM.

\subsection{Activation Extraction}
Given an input sequence of tokens \(x = (x_1,\dots,x_T)\), we extract post layer residual-stream hidden states across a set of \(L\) continuous transformer layers and apply a mask to exclude padding positions in all subsequent steps.\footnote{We also evaluated using only the final token representation; token-aggregated representations provided a more stable signal in our SAE setting}
Let \(h_{\ell,t} \in \mathbb{R}^{d}\) denote the hidden state at layer \(\ell \in \{1,\dots,L\}\) and token position \(t \in \{1,\dots,T\}\). This yields a layerwise collection of token activations \(\{h_{\ell,1},\dots,h_{\ell,T}\}_{\ell=1}^{L}\).

\subsection{Sparse Feature Decomposition with SAEs}
To obtain an interpretable, sparse representation, we pass each token activation \(h_{\ell,t}\) through a pre-trained layer-specific sparse autoencoder (SAE) encoder:
\begin{equation}
z_{\ell,t} = \mathrm{Enc}_{\ell}(h_{\ell,t}) \in \mathbb{R}^{D},
\end{equation}
where \(D \gg d\). The resulting vectors \(z_{\ell,t}\) are sparse, and their non-zero entries correspond to learned features that are approximately mono-semantic.

\textbf{Token-axis pooling.}
For each layer \(\ell\), we aggregate token-level SAE features into a single layer representation by averaging over token positions:
\begin{equation}
\bar{z}_{\ell} = \frac{1}{\sum_{t=1}^{T} m_t}\sum_{t=1}^{T} m_t \, z_{\ell,t},
\end{equation}
where \(m_t \in \{0,1\}\) is the \emph{padding mask}, 1 denoting non-padding token. This produces a pooled sparse feature trajectory \(\{\bar{z}_1,\dots,\bar{z}_L\}\).

\textbf{Global-density feature masking.}
Some SAE features are overly frequent and carry little domain-specific information. We therefore prune features using global activation density statistics (e.g., Neuronpedia-reported densities). Let \(\rho_{\ell,j}\) denote the global activation density of feature \(j\) at layer \(\ell\). We construct a binary mask \(g_{\ell} \in \{0,1\}^{D}\) that removes features above a density threshold \(\theta\):
\begin{equation}
g_{\ell,j} = \mathbbm{1}\left[\rho_{\ell,j} \le \theta\right],
\qquad
\tilde{z}_{\ell} = \bar{z}_{\ell} \odot g_{\ell},
\end{equation}
where \(\odot\) is element-wise multiplication.

\textbf{Top-\(k\) binarization.}
Finally, we convert \(\tilde{z}_{\ell}\) into a binary sparse distributed representation (SDR) by selecting the indices of the top-\(k\) remaining features by magnitude:
\begin{equation}
A_{\ell} = \mathrm{TopK}(\tilde{z}_{\ell}, k),
\qquad
s_{\ell}[j] = \mathbbm{1}[j \in A_{\ell}],
\end{equation}
yielding a binary sequence \(\{s_1,\dots,s_L\}\) used by downstream sequential scoring modules.


\subsection{Sequential Anomaly Scoring}
\label{sec:seq_scoring}

For an input $x$ we obtain a depthwise SDR sequence $\{s_\ell\}_{\ell=1}^{L}$, where $s_\ell\in\{0,1\}^D$ has exactly $k$ active bits.
Let $A_\ell(x)=\{j\in[D]\mid s_\ell[j]=1\}$ be the active set at layer $\ell$.
Each backend defines a per-layer anomaly $a_\ell(x)$ for $\ell=2,\dots,L$ and we aggregate
\begin{equation}
S(x)=\frac{1}{L-1}\sum_{\ell=2}^{L} a_\ell(x).
\end{equation}

\textbf{Markov (default).}
From ID data we count adjacent-layer co-activations
$C_\ell(i,j)=\#\{x\in\mathcal{D}_{ID}: i\in A_{\ell-1}(x),\, j\in A_\ell(x)\}$ and marginals $N_\ell(i)=\sum_j C_\ell(i,j)$.
With Laplace smoothing $\alpha$, we set
\begin{equation}
p_\ell(j\mid i)=\frac{C_\ell(i,j)+\alpha}{N_\ell(i)+\alpha D}.
\end{equation}
At test time we score transitions by the negative mean log-likelihood over active pairs:
\begin{equation}
a_\ell(x)= -\frac{1}{|A_{\ell-1}||A_\ell|}\sum_{i\in A_{\ell-1}}\sum_{j\in A_\ell}\log p_\ell(j\mid i),
\end{equation}
assigning unseen transitions the smoothed floor probability.

\textbf{HTM and RNN.}
As higher-order baselines, we also score sequences using (i) HTM temporal memory anomaly (fraction of active bits not predicted),
and (ii) an LSTM/GRU next-layer predictor trained on ID sequences with per-bit BCE loss; full details are in the appendix.

\section{Experiments}
\subsection{Experimental Setup}
We run all experiments on pretrained \textbf{Gemma2-2B} and \textbf{Gemma2-9B}.\cite{gemmateam2024gemma2improvingopen} From each transformer block we extract residual-stream activations and encode them using a pretrained \textbf{16k-dimensional sparse autoencoder (SAE)}; SAE activations are mean-pooled over tokens and converted to a binary sparse distributed representation (SDR) via Top-$k$ selection. Unless stated otherwise, we use $k{=}10$. We evaluate ID-only OOD detection under two regimes: \textbf{far-OOD}, where ID is 20 Newsgroups and OOD is drawn from SST-2, MNLI, RTE, IMDB, and CLINC150; and \textbf{near-OOD}, where ID and OOD share the same dataset but differ by semantic class, using AGNews (4 one-vs-all splits), ROSTD, SNIPS, and CLINC150 (see Table~\ref{tab:near_ood_main} for split details). Our detector operates on depth-wise SAE-SDR sequences using three alternative sequential backends: a sparse first-order \textbf{Markov} transition model, \textbf{HTM} \cite{Cui_2016}, and an \textbf{RNN} (LSTM) trained to predict next-layer activations with BCE; unless stated otherwise we use Markov. As a baseline we use the \textbf{likelihood-ratio (LR)} criterion between a base model and an in-domain fine-tuned model as it achieves SOTA \cite{zhang2025finetunedlargelanguagemodel}. We report AUROC, AUPR (OOD as positive), and FPR@95\%TPR, averaged over three seeds where applicable.

\subsection{Representation Analysis}
\subsubsection{domain characterization}
\begin{figure}[t]
  \centering
  \includegraphics[width=\linewidth]{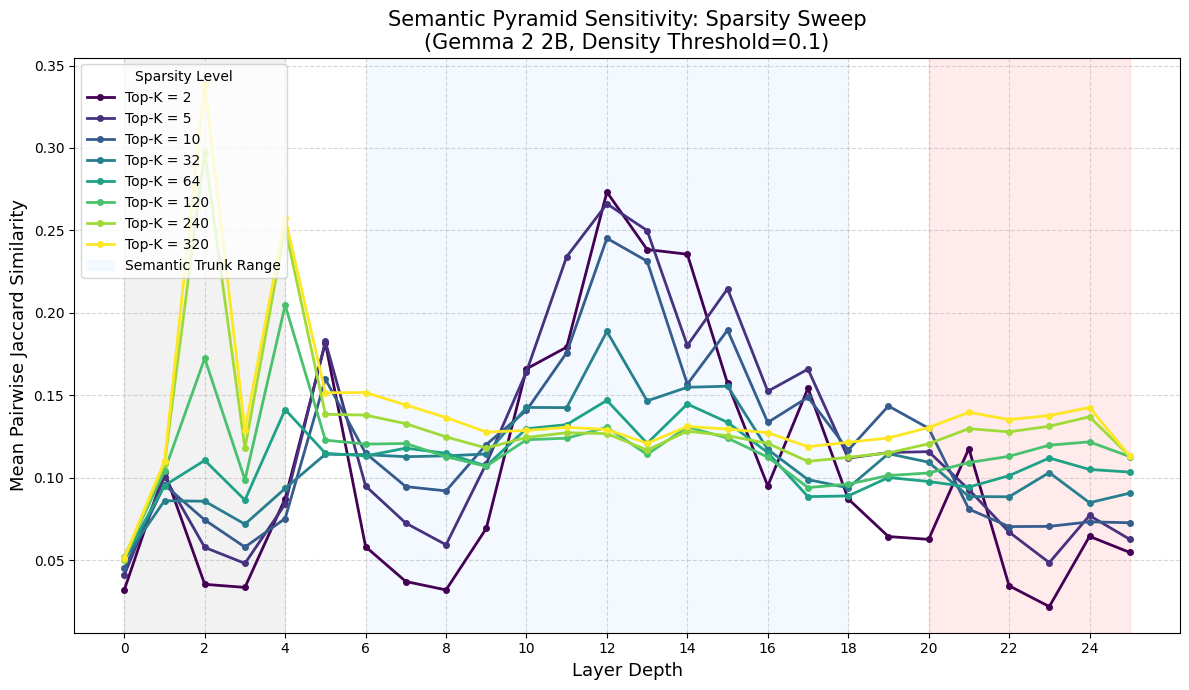}
  \caption{Mean Jaccard similarity across layers for different Top-\(k\) settings (computed on all pairs of a subset of 1000 in-domain samples).}
  \label{fig:pyramid_topk}
\end{figure}

We begin by characterizing where and how strongly a domain signal appears in the SAE feature space across depth. For an input \(x\), recall that our representation produces a masked pooled vector \(\tilde z_\ell(x)\) at each layer \(\ell\), and an active feature set
\begin{equation}
A_\ell(x) = \mathrm{TopK}(\tilde z_\ell(x), k).
\end{equation}
For a fixed \(k\), we measure \emph{within-domain consistency} at each layer by sampling pairs of in-domain inputs \((x,x')\) and computing the Jaccard similarity
\begin{equation}
J_\ell(x,x')=\frac{|A_\ell(x)\cap A_\ell(x')|}{|A_\ell(x)\cup A_\ell(x')|}.
\end{equation}
We report the mean of \(J_\ell\) over sampled pairs as a function of layer and sweep \(k\), shown in Figure~\ref{fig:pyramid_topk}.

Across \(k\), we observe a characteristic depth profile: early layers exhibit lower agreement in top activating features, mid-layers show increased similarity in top activating features, and late layers become more input-specific. This motivates selecting a moderate \(k\) that preserves stable shared structure while avoiding excessive noise (we use \(k{=}10\) in subsequent analyses unless stated otherwise). The peak around mid-layers suggests that domain-consistent features are most prominent in this region in most activating features, which motivates examining whether \emph{transitions} across layers carry additional domain structure.

\subsubsection{Dynamic analysis}
The static agreement analysis indicates that in-domain inputs share features in particular depth regions, but it does not reveal whether these features are domain specifics. We therefore analyze the specificity of the features across layers and transitions between them.

Given a domain \(D\), for each start layer \(\ell\) and hop length \(N\), we construct a registry of all \(N\)-hop feature tuples observed in the in-domain data:
\begin{equation}
\mathcal{V}_{\ell,N} \coloneqq
\left\{
(u_0,\ldots,u_N)\ \middle|\ 
\begin{aligned}
&\exists x \in D,\\
&\forall i = 0,\ldots,N:\ u_i \in A_{\ell+i}(x)
\end{aligned}
\right\}.
\end{equation}

We consider \(N \in \{0,1,2\}\), where \(N=0\) corresponds to a static registry.

\textbf{Trajectory-based scoring.}
For a sample \(x\), define the set of induced trajectories starting at layer \(l\) with hop length \(H\) as
\begin{equation}
\mathcal{T}_{l,H}(x)=A_l(x)\times A_{l+1}(x)\times \dots \times A_{l+H}(x),
\end{equation}
with \(|\mathcal{T}_{l,H}(x)|=\prod_{i=0}^{H}|A_{l+i}(x)|\).
We score how \emph{typical} these trajectories are under the in-domain registry by the fraction of induced tuples present in $\mathcal{T}_{l,H}(x)$:
\begin{equation}
\mathrm{Score}_H(x,l)=
\frac{\left|\mathcal{T}_{l,H}(x)\cap \mathcal{V}_{l,H}\right|}{|\mathcal{V}_{l,H}|}.
\end{equation}

\begin{figure}[t]
  \centering
  \includegraphics[width=\linewidth]{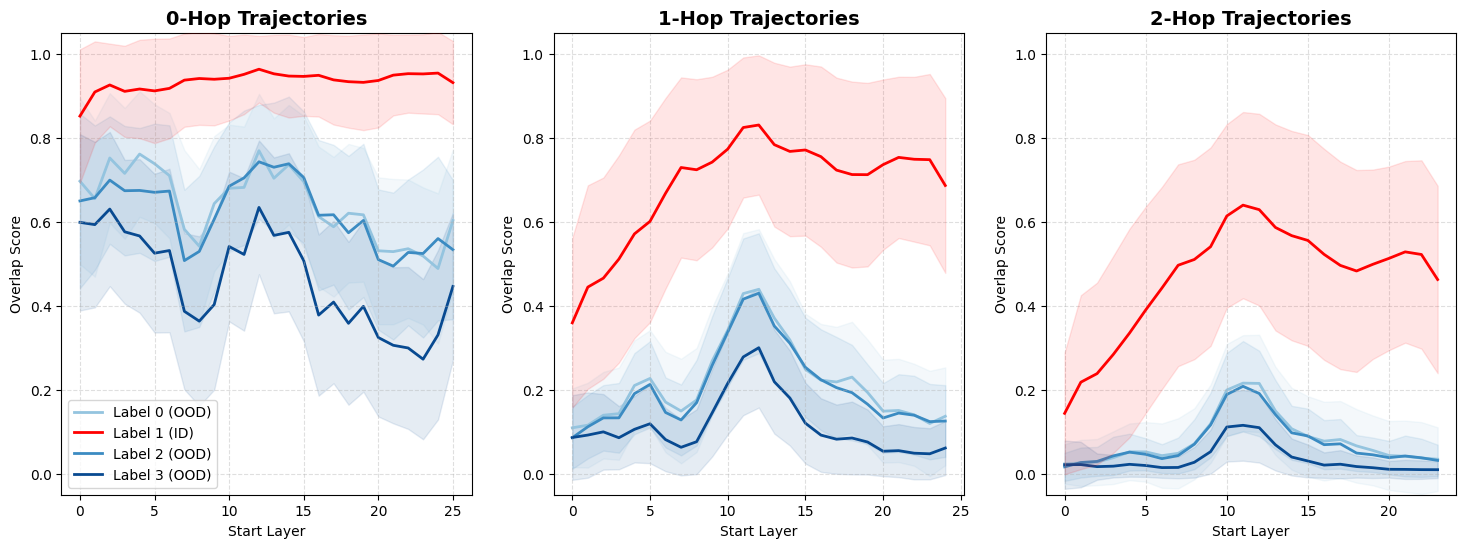}
  \caption{Layer-wise trajectory validity scores for hop lengths \(H\in\{0,1,2\}\) (Top-\(k\)=10). Lines show mean and shaded regions indicate \(\pm 1\) standard deviation.}
  \label{fig:traj_scores_4hop}
\end{figure}

Figure~\ref{fig:traj_scores_4hop} shows that (i) different semantic classes induce distinguishable depthwise trajectories, and (ii) longer hop lengths exhibit increased variance within a class, consistent with encoding more context-specific information. However, the registry-based construction is inherently coarse: it accumulates many incidental tuples, which increases noise and can lead to elevated FPR@95 as can be seen in \ref{tab:hop_metrics}. This motivates replacing the explicit registry with learned sequential scoring functions that emphasize recurring transition structure while down-weighting spurious co-activations.

\begin{table}[ht]
    \centering
    \caption{Mean OOD Detection Metrics across Trajectory Hop Lengths}
    \label{tab:hop_metrics}
    \begin{tabular}{@{}ccccc@{}}
        \toprule
        \textbf{Hop} & \textbf{AUROC} & \textbf{AUPR} & \textbf{FPR@95} & \textbf{Pairs Evaluated} \\ \midrule
        0            & 0.8445         & 0.8092        & 0.4451          & 12                      \\
        1            & 0.9159         & 0.9226        & 0.3902          & 12                      \\
        2            & 0.9128         & 0.9204        & 0.4054          & 12                      \\ \bottomrule
    \end{tabular}
\end{table}

\textbf{Takeaway.}
The static overlap results indicate that in-domain inputs share stable high activating feature structure in mid-depth layers, while the trajectory analysis shows that depthwise transitions carry additional domain-specific regularities. However, explicit tuple registries are coarse and accumulate incidental patterns, motivating a learned sequential scoring function that can emphasize recurring transition structure while suppressing spurious co-activations. In the ablation section, we evaluate sequential scoring backends trained only on in-scope data.

Unless stated otherwise, we report results using the first-order Markov backend; alternative backends (HTM and RNN) are compared in the ablation study.

\subsection{Main Results}
We evaluate scope gating as OOD detection under two regimes: \emph{far-OOD}, where out-of-scope data differs in both semantics and domain/covariates, and \emph{near-OOD}, where in-scope and out-of-scope inputs share similar style but differ semantically. We report AUROC, AUPR (OOD as positive), and FPR@95\%TPR.

\subsubsection{Baselines}
Our primary comparator is the likelihood-ratio (LR) criterion between a pretrained base LLM and an in-scope-adapted variant, which has been shown to be a strong OOD detector in an ID-only setting. \cite{zhang2025finetunedlargelanguagemodel}
This baseline is particularly relevant in our setting because it does not require labeled or OOD data and naturally fits the scope-gating interpretation of OOD detection.

\subsubsection{Far-OOD Results}
In the far-OOD setting, the in-scope dataset 20NG is fixed and multiple out-of-scope datasets are used as OOD sources. This evaluates coarse rejection under substantial domain/covariate shift. 
\begin{table*}[t]
\centering
\small
\setlength{\tabcolsep}{4pt}
\renewcommand{\arraystretch}{1.2}
\caption{Far-OOD results (ID-only training).}
\begin{tabular}{l l l c c c c}
\toprule
\textbf{ID} & \textbf{OOD} & \textbf{Method} &
{\textbf{AUROC} $\uparrow$} & {\textbf{AUPR (OOD)} $\uparrow$} & {\textbf{FPR95} $\downarrow$} \\
\midrule

\multirow{4}{*}{20NG} & \multirow{4}{*}{SST-2} &
Gemma2-2b + 16k\_SAE + Markov & 0.9782 & 0.7683 & 0.1101  \\
& & Gemma2-9b + 16k\_SAE + Markov & 0.9866 & 0.8713  & 0.0665 \\
& & Gemma2-2b LR baseline & \textbf{0.9988} & \textbf{0.9871} & \textbf{0.0027}  \\
& & Gemma2-9b LR baseline & 0.9980 & 0.9994 & 0.004  \\
\midrule

\multirow{4}{*}{20NG} & \multirow{4}{*}{MNLI} &
Gemma2-2b + 16k\_SAE + Markov & 0.9699 & 0.9680 & 0.1984  \\
& & Gemma2-9b + 16k\_SAE + Markov & 0.9763  & 0.9768  & 0.1628  \\
& & Gemma2-2b LR baseline & \textbf{0.9989} & \textbf{0.9988} & \textbf{0.0027}  \\
& & Gemma2-9b LR baseline & 0.9982 & 0.997 & 0.003 \\
\midrule

\multirow{4}{*}{20NG} & \multirow{4}{*}{RTE} &
Gemma2-2b + 16k\_SAE + Markov & 0.9850 & 0.6422 & 0.0325 \\
& & Gemma2-9b + 16k\_SAE + Markov & 0.9884  & 0.7333  & 0.0469 \\
& & Gemma2-2b LR baseline & \textbf{0.9991} & 0.9950 & \textbf{0.0022}  \\
& & Gemma2-9b LR baseline & 0.9987 & \textbf{0.9994} & 0.003 \\
\midrule

\multirow{4}{*}{20NG} & \multirow{4}{*}{IMDB} &
Gemma2-2b + 16k\_SAE + Markov & 0.9171 & 0.9058 & 0.6243 \\
& & Gemma2-9b + 16k\_SAE + Markov & 0.9540  & 0.9544 & 0.3361 \\
& & Gemma2-2b LR baseline & \textbf{0.9978} & \textbf{0.9992} & \textbf{0.0072}  \\
& & Gemma2-9b LR baseline & 0.9945 & 0.9840 & 0.017 \\
\bottomrule
\end{tabular}

\label{tab:far_ood_main}
\end{table*}

\subsubsection{Near-OOD Results}
In the near-OOD setting, in-scope and out-of-scope examples are drawn from the same dataset split by semantic classes, preserving similar surface form while changing intent/topic. This setting stresses fine-grained semantic discrimination, and is commonly used to evaluate scope detection.

\begin{table*}[t]
\centering
\small
\setlength{\tabcolsep}{6pt} 
\renewcommand{\arraystretch}{1.2}
\caption{
Near-OOD results (ID-only training).
Splits: \textbf{AGNews} uses 4 one-vs-all runs (reported as $c_3/c_2/c_1/c_0$ in each metric cell).
\textbf{ROSTD} uses the dataset’s predefined in-domain classes (13) vs. its OOD set.
\textbf{SNIPS} uses ID intents \{AddToPlaylist, PlayMusic, RateBook, SearchCreativeWork, SearchScreeningEvent\} and OOD intents \{GetWeather, BookRestaurant\}.
\textbf{CLINC150} uses the dataset’s predefined in-domain classes (150) vs. its OOD set.
}
\begin{tabularx}{\textwidth}{
l
@{\hspace{10pt}}  
>{\raggedright\arraybackslash}X
@{\hspace{12pt}}  
c c c c
}
\toprule
\textbf{Dataset} & \textbf{Method} &
{\textbf{AUROC} $\uparrow$} & {\textbf{AUPR (OOD)} $\uparrow$} & {\textbf{FPR95} $\downarrow$} \\
\midrule

\multirow{4}{*}{AGNews} &
Gemma2-2b + 16k\_SAE + Markov &
\textbf{0.90}  /\textbf{0.98}  /\textbf{0.92}  /\textbf{0.88} &
\textbf{0.96  /0.99  /0.97  /0.95} &
0.50  \textbf{/0.03  /0.41  /0.51} \\
& Gemma2-9b + 16k\_SAE + Markov &
0.87  /0.90  /0.83  /0.81 &
0.95  /0.96  /0.93  /0.92 &
\textbf{0.45}  /0.31  /0.54  /0.59 \\
& Gemma2-2b LR baseline &
0.75  /0.76  /0.82  /0.73 &
0.89  /0.90  /0.93  /0.88 &
0.76  /0.68  /0.67  /0.80 \\
& Gemma2-9b LR baseline &
0.73  /0.74  /0.79  /0.75 &
0.89  /0.86  /0.90  /0.87 &
0.79  /0.77  /0.70  /0.75 \\
\midrule

\multirow{4}{*}{ROSTD} &
Gemma2-2b + 16k\_SAE + Markov & 0.9532 & \textbf{0.9548} & \textbf{0.1757} \\
& Gemma2-9b + 16k\_SAE + Markov & 0.9445 & 0.9454 & 0.2068 \\
& Gemma2-2b LR baseline & \textbf{0.9726} & 0.9455 & 0.1690 \\
& Gemma2-9b LR baseline & 0.9549 & 0.9165 & 0.2597 \\
\midrule

\multirow{4}{*}{SNIPS} &
Gemma2-2b + 16k\_SAE + Markov & 0.9699  & 0.9338 & 0.1525 \\
& Gemma2-9b + 16k\_SAE + Markov & 0.9001   & 0.8287 & 0.3150 \\
& Gemma2-2b LR baseline & 0.9746 & 0.9423 & 0.1528 \\
& Gemma2-9b LR baseline & \textbf{0.9798}  & \textbf{0.9546}  & \textbf{0.0873} \\
\midrule

\multirow{4}{*}{CLINC150} &
Gemma2-2b + 16k\_SAE + Markov & 0.7387  & 0.0127*  & 0.8333 \\
& Gemma2-9b + 16k\_SAE + Markov & 0.8364 & \textbf{0.7278}  & \textbf{0.4825} \\
& Gemma2-2b LR baseline & \textbf{0.8661} & 0.6449 & 0.5607  \\
& Gemma2-9b LR baseline & 0.8494 & 0.6296 & 0.6440  \\

\bottomrule
\end{tabularx}
\label{tab:near_ood_main}
\end{table*}
\textbf{Discussion.}
Near-OOD detection is substantially more challenging than far-OOD, as in- and out-of-scope examples share surface form and differ only in fine-grained semantics.
On SNIPS, ROSTD and AGNews our method is competitive with the LR baseline and achieves comparable FPR95, indicating that transition regularities in internal representations can capture intent-level shifts when the underlying domains are semantically coherent and sufficiently coarse.

On CLINC150, however, performance degrades sharply for the 2B model, most notably in AUPR.
We attribute this failure to a \emph{representation resolution mismatch} rather than a modeling artifact.
CLINC150 defines 150 highly specific intents spanning very different tasks (e.g., mechanical instructions, banking queries, authentication issues, customer support, etc.), while our method operates on a finite set of SAE features whose semantic granularity is necessarily coarser.

As a result, many distinct intents collapse onto overlapping subsets of SAE features and therefore induce highly similar, high-entropy trajectories across depth touching upon many abstract concepts.
In this regime, different intents are not separated by stable, low-entropy manifolds in representation space, but instead share large portions of their activation patterns.
Since the sequential scorer relies on consistent inter-layer transition structure to distinguish in-domain from out-of-domain behavior, this representational aliasing makes reliable discrimination impossible.
The stronger performance of the 9B model suggests that increased representational capacity partially alleviates this bottleneck, but does not remove it entirely.
We therefore view CLINC150 as illustrating a fundamental limitation of representation-dynamics based domain restriction: when the task taxonomy is significantly finer-grained than the resolution of the underlying features, domain boundaries cease to be well-defined in internal representation space.

\subsection{Case Study: Symmetric ``Hard-OOD'' Boundary Examples in AGNews}
\label{sec:case_study_agnews}

Near-OOD settings can contain genuine semantic overlap between the ID and OOD classes, making some OOD inputs effectively \emph{in-domain in intent}.
To illustrate what our detector learns beyond dataset labels, we conduct a focused case study on AGNews labels \textbf{Business (2)} and \textbf{Sci/Tech (3)} using the Markov backend from \cref{sec:seq_scoring}.
For each direction, we train the Markov transition model on ID only and then identify \emph{hard OOD} examples: OOD test inputs with the \emph{lowest} anomaly score $S(x)$ (i.e., those most confidently judged as ID by the transition model).
Concretely, hard OOD corresponds to false negatives under the OOD-as-positive convention.

\textbf{Observation: hard OOD is semantically ``in-domain'' in both directions.}
When training on \textbf{Business} and testing on \textbf{Sci/Tech} as OOD, hard-OOD Sci/Tech articles are predominantly \emph{finance/earnings/market} stories (often about tech companies), and thus resemble Business in content.
Conversely, when training on \textbf{Sci/Tech} and treating \textbf{Business} as OOD, hard-OOD Business articles frequently describe \emph{telecom infrastructure, software releases, and product launches}, resembling Sci/Tech.
This symmetry indicates that many apparent “errors” arise from \emph{true boundary overlap} rather than arbitrary model failure.

\textbf{SAE-transition evidence: hard OOD follows ID-like internal dynamics.}
Beyond the raw prompt text, we inspect which SAE feature transitions explain the low anomaly.
For each hard-OOD example, we extract the top contributing active feature-pairs $(i\!\rightarrow\!j)$ across depth (highest $-\log p_\ell(j\mid i)$ among active pairs; see \cref{sec:seq_scoring}).
We decode feature indices to human-readable labels via Neuronpedia.

The resulting explanations show that hard OOD tends to activate and transition through features whose semantics match the ID domain signature (e.g., market/earnings concepts under Business-ID; network/software concepts under Sci/Tech-ID), supporting the hypothesis that the Markov score tracks \emph{domain-consistent internal trajectories} rather than superficial cues.

\begin{figure}[t]
\centering
\begin{minipage}{0.96\linewidth}
\small
\textbf{Business-ID (2) $\rightarrow$ hard OOD from Sci/Tech (3)}\\
\emph{``Gateway Reports Smaller Quarterly Loss \ldots continues to restructure \ldots integrate its acquisition \ldots''}\\[0.5em]
\textbf{Decoded transition evidence (top contributors):}
\begin{itemize}
    \item 16$\rightarrow$17: \textit{terms related to business acquisitions and financial transactions}
    $\rightarrow$ \textit{references to significant events and metrics in financial contexts}
    \item 18$\rightarrow$19: \textit{terms related to acquisitions and financial transactions}
    $\rightarrow$ \textit{references to financial data and reporting metrics}
\end{itemize}
\end{minipage}
\caption{\small Example of a hard OOD input from Sci/Tech that appears Business-like under the Markov score.}
\end{figure}

\begin{figure}[t]
\centering
\begin{minipage}{0.96\linewidth}
\small
\textbf{Business-ID (2) $\rightarrow$ hard OOD from Sci/Tech (3)}\\
\emph{``Intel Posts Higher Profit, Sales \ldots earnings \ldots sales \ldots demand \ldots''}\\[0.5em]
\textbf{Decoded transition evidence (top contributors):}
\begin{itemize}
    \item 18$\rightarrow$19: \textit{phrases pertaining to economic trends and conditions}
    $\rightarrow$ \textit{references to trade agreements and restrictions}
    \item 23$\rightarrow$24: \textit{phrases related to financial performance and changes in revenue}
    $\rightarrow$ \textit{references to organized meetings or discussions related to trade and technology policies}
\end{itemize}
\end{minipage}
\caption{\small Another hard OOD example under Business-ID.}
\end{figure}

\begin{figure}[t]
\centering
\begin{minipage}{0.96\linewidth}
\small
\textbf{Sci/Tech-ID (3) $\rightarrow$ hard OOD from Business (2)}\\
\emph{``Microsoft Readies Next Business IM Server \ldots enterprise instant messaging software \ldots''}\\[0.5em]
\textbf{Decoded transition evidence (top contributors):}
\begin{itemize}
    \item 16$\rightarrow$17: \textit{information related to software updates and their details}
    $\rightarrow$ \textit{terms related to electron interactions and superconductivity}
    \item 16$\rightarrow$17: \textit{descriptions and features of new technology products}
    $\rightarrow$ \textit{terms related to electron interactions and superconductivity}
\end{itemize}
\end{minipage}
\caption{\small Hard OOD example under Sci/Tech-ID that appears Sci/Tech-like.}
\end{figure}

\begin{figure}[t]
\centering
\begin{minipage}{0.96\linewidth}
\small
\textbf{Sci/Tech-ID (3) $\rightarrow$ hard OOD from Business (2)}\\
\emph{``Cingular to upgrade wireless data network \ldots handle high-speed data transmissions \ldots''}\\[0.5em]
\textbf{Decoded transition evidence (top contributors):}
\begin{itemize}
    \item 17$\rightarrow$18: \textit{references to financial institutions and related operational terms}
    $\rightarrow$ \textit{references to ``data'' and its contexts in research and analysis}
\end{itemize}
\end{minipage}
\caption{\small Another hard OOD example under Sci/Tech-ID.}
\end{figure}

\textbf{Takeaway.}
Across both directions, the \emph{lowest-anomaly OOD} examples are precisely those whose \emph{content} overlaps the ID domain (e.g., earnings-driven tech reporting or telecom/software business news), and their \emph{internal SAE trajectories} expose corresponding domain-aligned feature transitions.
This supports the interpretation that our method performs \emph{semantic scope gating} by modeling consistent depthwise feature dynamics, and that many ``mistakes'' are explainable boundary cases rather than spurious failures.

\subsection{Ablations}
\subsubsection{Sequential backend ablation}

\textbf{Setup.}
We compare three sequential backends operating on the same SAE-derived binary SDR sequences across transformer depth: (i) \textbf{HTM}, a high-order temporal memory model; (ii) \textbf{RNN}, an LSTM next-layer predictor trained with a BCE objective; and (iii) \textbf{Markov}, a first-order transition model estimating sparse adjacent-layer transition probabilities. All models are trained on in-distribution (ID) data only and evaluated for OOD detection using AUROC, AUPR, and FPR@95 (lower is better).

\textbf{Aggregated results.}
To summarize the four ID-class runs, we report the mean performance across runs in Table~\ref{tab:seq_ablation_mean}.

\begin{table}[h]
\centering
\caption{Mean OOD detection performance across four ID-class runs for different sequential backends.}
\begin{tabular}{lccc}
\hline
\textbf{Method} & \textbf{AUROC} $\uparrow$ & \textbf{AUPR} $\uparrow$ & \textbf{FPR@95} $\downarrow$ \\
\hline
HTM    & 0.912 & 0.963 & 0.384 \\
Markov & \textbf{0.919} & \textbf{0.968} & \textbf{0.371} \\
RNN    & 0.910 & 0.965 & 0.392 \\
\hline
\end{tabular}

\label{tab:seq_ablation_mean}
\end{table}

\textbf{Interpretation.}
A simple first-order Markov model matches or slightly outperforms both high-order backends (HTM and RNN) on average. This indicates that, under the current SAE-SDR representation, \textbf{most of the discriminative signal for OOD detection is already present in adjacent-layer transition statistics}. Higher-order sequence modeling provides no consistent advantage, suggesting that transformer depth dynamics in this setting are close to first-order Markovian for the purpose of ID/OOD separation.

\subsubsection{SAE Ablation}
We try to extend our method on the raw internal of the transformer with similar setting to our setup, we binarized using the strongest 32 activations to the raw embeddings we utilize the HTM sequential scorer to model the transitions. 
we report the mean performance for all ag\_news splits. Learning on raw internals show bad results, which motivates the use of SAE to disentangle the superposition phenomenon of neurons that might explain such results.
\begin{table}[h]
\centering
\caption{Mean OOD detection performance across four ID-class runs on the raw internals.}
\begin{tabular}{lccc}
\hline
\textbf{Method} & \textbf{AUROC} $\uparrow$ & \textbf{FPR@95} $\downarrow$ \\
\hline
HTM + raw\_activations    & 0.753 & 0.638 \\
\hline
\end{tabular}
\label{tab:raw_activation_ablation}
\end{table}

\section{Future Work and Discussion}
This paper focuses on binarized SAE vectors. A natural next step is to extend the approach to incorporate activation strengths as well. In addition, our experiments are limited to SAE-16k; it would be interesting to evaluate more expressive SAEs (e.g., 65k) in future work.

We show that \emph{depthwise transition regularities} in sparse, interpretable SAE feature space provide a strong ID-only signal for scope gating, especially in far-OOD and several near-OOD settings. A key limitation emerges when the task taxonomy is substantially finer-grained than the effective resolution of SAE features (e.g., CLINC150 for Gemma2-2B): representational aliasing can yield similar trajectories across distinct intents, degrading discrimination.

\section*{Impact Statement}
This work targets safer and more reliable deployment of LLM-based systems by enabling lightweight \emph{scope gating} that can reject or reroute out-of-domain requests using only in-domain data.
Potential risks include uneven false-reject rates across user groups or topics, and false-accepts that allow out-of-scope behavior; both can harm user experience or system safety.
Mitigations include careful threshold calibration on representative in-domain traffic, monitoring error patterns over time, and using the gate as one component in a defense-in-depth design (e.g., combined with policy filters and human review for high-stakes actions).

\bibliography{example_paper}

\begin{thebibliography}{33}
\providecommand{\natexlab}[1]{#1}
\providecommand{\url}[1]{\texttt{#1}}
\expandafter\ifx\csname urlstyle\endcsname\relax
  \providecommand{\doi}[1]{doi: #1}\else
  \providecommand{\doi}{doi: \begingroup \urlstyle{rm}\Url}\fi

\bibitem[Arad et~al.(2025)Arad, Mueller, and Belinkov]{Arad_2025}
Arad, D., Mueller, A., and Belinkov, Y.
\newblock Saes are good for steering – if you select the right features.
\newblock In \emph{Proceedings of the 2025 Conference on Empirical Methods in Natural Language Processing}, pp.\  10252–10270. Association for Computational Linguistics, 2025.
\newblock \doi{10.18653/v1/2025.emnlp-main.519}.
\newblock URL \url{http://dx.doi.org/10.18653/v1/2025.emnlp-main.519}.

\bibitem[Bloom et~al.(2024)Bloom, Tigges, Duong, and Chanin]{bloom2024saetrainingcodebase}
Bloom, J., Tigges, C., Duong, A., and Chanin, D.
\newblock Saelens.
\newblock \url{https://github.com/decoderesearch/SAELens}, 2024.

\bibitem[Chen et~al.(2023)Chen, Zaharia, and Zou]{chen2023frugalgptuselargelanguage}
Chen, L., Zaharia, M., and Zou, J.
\newblock Frugalgpt: How to use large language models while reducing cost and improving performance, 2023.
\newblock URL \url{https://arxiv.org/abs/2305.05176}.

\bibitem[Chen et~al.(2022)Chen, Bi, Gao, and Sun]{chen2022holisticsentenceembeddingsbetter}
Chen, S., Bi, X., Gao, R., and Sun, X.
\newblock Holistic sentence embeddings for better out-of-distribution detection, 2022.
\newblock URL \url{https://arxiv.org/abs/2210.07485}.

\bibitem[Cui et~al.(2016)Cui, Ahmad, and Hawkins]{Cui_2016}
Cui, Y., Ahmad, S., and Hawkins, J.
\newblock Continuous online sequence learning with an unsupervised neural network model.
\newblock \emph{Neural Computation}, 28\penalty0 (11):\penalty0 2474–2504, November 2016.
\newblock ISSN 1530-888X.
\newblock \doi{10.1162/neco_a_00893}.
\newblock URL \url{http://dx.doi.org/10.1162/NECO_a_00893}.

\bibitem[Cunningham et~al.(2023)Cunningham, Ewart, Riggs, Huben, and Sharkey]{cunningham2023sparseautoencodershighlyinterpretable}
Cunningham, H., Ewart, A., Riggs, L., Huben, R., and Sharkey, L.
\newblock Sparse autoencoders find highly interpretable features in language models, 2023.
\newblock URL \url{https://arxiv.org/abs/2309.08600}.

\bibitem[Devlin et~al.(2019)Devlin, Chang, Lee, and Toutanova]{devlin2019bertpretrainingdeepbidirectional}
Devlin, J., Chang, M.-W., Lee, K., and Toutanova, K.
\newblock Bert: Pre-training of deep bidirectional transformers for language understanding, 2019.
\newblock URL \url{https://arxiv.org/abs/1810.04805}.

\bibitem[Elhage et~al.(2022)Elhage, Hume, Olsson, Schiefer, Henighan, Kravec, Hatfield-Dodds, Lasenby, Drain, Chen, Grosse, McCandlish, Kaplan, Amodei, Wattenberg, and Olah]{elhage2022toymodelssuperposition}
Elhage, N., Hume, T., Olsson, C., Schiefer, N., Henighan, T., Kravec, S., Hatfield-Dodds, Z., Lasenby, R., Drain, D., Chen, C., Grosse, R., McCandlish, S., Kaplan, J., Amodei, D., Wattenberg, M., and Olah, C.
\newblock Toy models of superposition, 2022.
\newblock URL \url{https://arxiv.org/abs/2209.10652}.

\bibitem[Hendrycks \& Gimpel(2018)Hendrycks and Gimpel]{hendrycks2018baselinedetectingmisclassifiedoutofdistribution}
Hendrycks, D. and Gimpel, K.
\newblock A baseline for detecting misclassified and out-of-distribution examples in neural networks, 2018.
\newblock URL \url{https://arxiv.org/abs/1610.02136}.

\bibitem[Hendrycks et~al.(2020)Hendrycks, Liu, Wallace, Dziedzic, Krishnan, and Song]{hendrycks2020pretrainedtransformersimproveoutofdistribution}
Hendrycks, D., Liu, X., Wallace, E., Dziedzic, A., Krishnan, R., and Song, D.
\newblock Pretrained transformers improve out-of-distribution robustness, 2020.
\newblock URL \url{https://arxiv.org/abs/2004.06100}.

\bibitem[Hu et~al.(2024)Hu, Bieker, Li, Jiang, Keigwin, Ranganath, Keutzer, and Upadhyay]{hu2024routerbenchbenchmarkmultillmrouting}
Hu, Q.~J., Bieker, J., Li, X., Jiang, N., Keigwin, B., Ranganath, G., Keutzer, K., and Upadhyay, S.~K.
\newblock Routerbench: A benchmark for multi-llm routing system, 2024.
\newblock URL \url{https://arxiv.org/abs/2403.12031}.

\bibitem[Huang et~al.(2025)Huang, Yu, Ma, Zhong, Feng, Wang, Chen, Peng, Feng, Qin, and Liu]{Huang_2025}
Huang, L., Yu, W., Ma, W., Zhong, W., Feng, Z., Wang, H., Chen, Q., Peng, W., Feng, X., Qin, B., and Liu, T.
\newblock A survey on hallucination in large language models: Principles, taxonomy, challenges, and open questions.
\newblock \emph{ACM Transactions on Information Systems}, 43\penalty0 (2):\penalty0 1–55, January 2025.
\newblock ISSN 1558-2868.
\newblock \doi{10.1145/3703155}.
\newblock URL \url{http://dx.doi.org/10.1145/3703155}.

\bibitem[Lang et~al.(2023)Lang, Zheng, Li, Sun, Huang, and Li]{lang2023surveyoutofdistributiondetectionnlp}
Lang, H., Zheng, Y., Li, Y., Sun, J., Huang, F., and Li, Y.
\newblock A survey on out-of-distribution detection in nlp, 2023.
\newblock URL \url{https://arxiv.org/abs/2305.03236}.

\bibitem[Lee et~al.(2018)Lee, Lee, Lee, and Shin]{lee2018simpleunifiedframeworkdetecting}
Lee, K., Lee, K., Lee, H., and Shin, J.
\newblock A simple unified framework for detecting out-of-distribution samples and adversarial attacks, 2018.
\newblock URL \url{https://arxiv.org/abs/1807.03888}.

\bibitem[Liang et~al.(2020)Liang, Li, and Srikant]{liang2020enhancingreliabilityoutofdistributionimage}
Liang, S., Li, Y., and Srikant, R.
\newblock Enhancing the reliability of out-of-distribution image detection in neural networks, 2020.
\newblock URL \url{https://arxiv.org/abs/1706.02690}.

\bibitem[Lin(2023)]{neuronpedia}
Lin, J.
\newblock Neuronpedia: Interactive reference and tooling for analyzing neural networks, 2023.
\newblock URL \url{https://www.neuronpedia.org}.
\newblock Software available from neuronpedia.org.

\bibitem[Lindsey et~al.(2024)Lindsey, Templeton, Marcus, Conerly, Batson, and Olah]{lindsey2024crosscoders}
Lindsey, J., Templeton, A., Marcus, J., Conerly, T., Batson, J., and Olah, C.
\newblock Sparse crosscoders for cross-layer features and model diffing.
\newblock \url{https://transformer-circuits.pub/2024/crosscoders/index.html}, 2024.
\newblock Accessed: 2026-01-29.

\bibitem[Liu et~al.(2025)Liu, Yu, Zhang, Xu, Lei, Lai, Gu, Ding, Men, Yang, Zhang, Deng, Zeng, Du, Zhang, Shen, Zhang, Su, Sun, Huang, Dong, and Tang]{liu2025agentbenchevaluatingllmsagents}
Liu, X., Yu, H., Zhang, H., Xu, Y., Lei, X., Lai, H., Gu, Y., Ding, H., Men, K., Yang, K., Zhang, S., Deng, X., Zeng, A., Du, Z., Zhang, C., Shen, S., Zhang, T., Su, Y., Sun, H., Huang, M., Dong, Y., and Tang, J.
\newblock Agentbench: Evaluating llms as agents, 2025.
\newblock URL \url{https://arxiv.org/abs/2308.03688}.

\bibitem[Liu et~al.(2024)Liu, Yao, Ton, Zhang, Guo, Cheng, Klochkov, Taufiq, and Li]{liu2024trustworthyllmssurveyguideline}
Liu, Y., Yao, Y., Ton, J.-F., Zhang, X., Guo, R., Cheng, H., Klochkov, Y., Taufiq, M.~F., and Li, H.
\newblock Trustworthy llms: a survey and guideline for evaluating large language models' alignment, 2024.
\newblock URL \url{https://arxiv.org/abs/2308.05374}.

\bibitem[Lu et~al.(2025)Lu, Wang, Sheng, He, Zheng, and Liang]{lu2025outofdistributiondetectiontaskorientedsurvey}
Lu, S., Wang, Y., Sheng, L., He, L., Zheng, A., and Liang, J.
\newblock Out-of-distribution detection: A task-oriented survey of recent advances, 2025.
\newblock URL \url{https://arxiv.org/abs/2409.11884}.

\bibitem[Marks et~al.(2025)Marks, Rager, Michaud, Belinkov, Bau, and Mueller]{marks2025sparsefeaturecircuitsdiscovering}
Marks, S., Rager, C., Michaud, E.~J., Belinkov, Y., Bau, D., and Mueller, A.
\newblock Sparse feature circuits: Discovering and editing interpretable causal graphs in language models, 2025.
\newblock URL \url{https://arxiv.org/abs/2403.19647}.

\bibitem[Paulo et~al.(2025)Paulo, Mallen, Juang, and Belrose]{paulo2025automaticallyinterpretingmillionsfeatures}
Paulo, G., Mallen, A., Juang, C., and Belrose, N.
\newblock Automatically interpreting millions of features in large language models, 2025.
\newblock URL \url{https://arxiv.org/abs/2410.13928}.

\bibitem[Podolskiy et~al.(2022)Podolskiy, Lipin, Bout, Artemova, and Piontkovskaya]{podolskiy2022revisitingmahalanobisdistancetransformerbased}
Podolskiy, A., Lipin, D., Bout, A., Artemova, E., and Piontkovskaya, I.
\newblock Revisiting mahalanobis distance for transformer-based out-of-domain detection, 2022.
\newblock URL \url{https://arxiv.org/abs/2101.03778}.

\bibitem[Räuker et~al.(2023)Räuker, Ho, Casper, and Hadfield-Menell]{räuker2023transparentaisurveyinterpreting}
Räuker, T., Ho, A., Casper, S., and Hadfield-Menell, D.
\newblock Toward transparent ai: A survey on interpreting the inner structures of deep neural networks, 2023.
\newblock URL \url{https://arxiv.org/abs/2207.13243}.

\bibitem[Schick et~al.(2023)Schick, Dwivedi-Yu, Dessì, Raileanu, Lomeli, Zettlemoyer, Cancedda, and Scialom]{schick2023toolformerlanguagemodelsteach}
Schick, T., Dwivedi-Yu, J., Dessì, R., Raileanu, R., Lomeli, M., Zettlemoyer, L., Cancedda, N., and Scialom, T.
\newblock Toolformer: Language models can teach themselves to use tools, 2023.
\newblock URL \url{https://arxiv.org/abs/2302.04761}.

\bibitem[Shazeer et~al.(2017)Shazeer, Mirhoseini, Maziarz, Davis, Le, Hinton, and Dean]{shazeer2017outrageouslylargeneuralnetworks}
Shazeer, N., Mirhoseini, A., Maziarz, K., Davis, A., Le, Q., Hinton, G., and Dean, J.
\newblock Outrageously large neural networks: The sparsely-gated mixture-of-experts layer, 2017.
\newblock URL \url{https://arxiv.org/abs/1701.06538}.

\bibitem[Sun et~al.(2022)Sun, Ming, Zhu, and Li]{sun2022outofdistributiondetectiondeepnearest}
Sun, Y., Ming, Y., Zhu, X., and Li, Y.
\newblock Out-of-distribution detection with deep nearest neighbors, 2022.
\newblock URL \url{https://arxiv.org/abs/2204.06507}.

\bibitem[Team et~al.(2024)Team, Riviere, Pathak, Sessa, Hardin, Bhupatiraju, Hussenot, Mesnard, Shahriari, Ramé, Ferret, Liu, Tafti, Friesen, Casbon, Ramos, Kumar, Lan, Jerome, Tsitsulin, Vieillard, Stanczyk, Girgin, Momchev, Hoffman, Thakoor, Grill, Neyshabur, Bachem, Walton, Severyn, Parrish, Ahmad, Hutchison, Abdagic, Carl, Shen, Brock, Coenen, Laforge, Paterson, Bastian, Piot, Wu, Royal, Chen, Kumar, Perry, Welty, Choquette-Choo, Sinopalnikov, Weinberger, Vijaykumar, Rogozińska, Herbison, Bandy, Wang, Noland, Moreira, Senter, Eltyshev, Visin, Rasskin, Wei, Cameron, Martins, Hashemi, Klimczak-Plucińska, Batra, Dhand, Nardini, Mein, Zhou, Svensson, Stanway, Chan, Zhou, Carrasqueira, Iljazi, Becker, Fernandez, van Amersfoort, Gordon, Lipschultz, Newlan, yeong Ji, Mohamed, Badola, Black, Millican, McDonell, Nguyen, Sodhia, Greene, Sjoesund, Usui, Sifre, Heuermann, Lago, McNealus, Soares, Kilpatrick, Dixon, Martins, Reid, Singh, Iverson, Görner, Velloso, Wirth, Davidow, Miller, Rahtz, Watson, Risdal,
  Kazemi, Moynihan, Zhang, Kahng, Park, Rahman, Khatwani, Dao, Bardoliwalla, Devanathan, Dumai, Chauhan, Wahltinez, Botarda, Barnes, Barham, Michel, Jin, Georgiev, Culliton, Kuppala, Comanescu, Merhej, Jana, Rokni, Agarwal, Mullins, Saadat, Carthy, Cogan, Perrin, Arnold, Krause, Dai, Garg, Sheth, Ronstrom, Chan, Jordan, Yu, Eccles, Hennigan, Kocisky, Doshi, Jain, Yadav, Meshram, Dharmadhikari, Barkley, Wei, Ye, Han, Kwon, Xu, Shen, Gong, Wei, Cotruta, Kirk, Rao, Giang, Peran, Warkentin, Collins, Barral, Ghahramani, Hadsell, Sculley, Banks, Dragan, Petrov, Vinyals, Dean, Hassabis, Kavukcuoglu, Farabet, Buchatskaya, Borgeaud, Fiedel, Joulin, Kenealy, Dadashi, and Andreev]{gemmateam2024gemma2improvingopen}
Team, G., Riviere, M., Pathak, S., Sessa, P.~G., Hardin, C., Bhupatiraju, S., Hussenot, L., Mesnard, T., Shahriari, B., Ramé, A., Ferret, J., Liu, P., Tafti, P., Friesen, A., Casbon, M., Ramos, S., Kumar, R., Lan, C.~L., Jerome, S., Tsitsulin, A., Vieillard, N., Stanczyk, P., Girgin, S., Momchev, N., Hoffman, M., Thakoor, S., Grill, J.-B., Neyshabur, B., Bachem, O., Walton, A., Severyn, A., Parrish, A., Ahmad, A., Hutchison, A., Abdagic, A., Carl, A., Shen, A., Brock, A., Coenen, A., Laforge, A., Paterson, A., Bastian, B., Piot, B., Wu, B., Royal, B., Chen, C., Kumar, C., Perry, C., Welty, C., Choquette-Choo, C.~A., Sinopalnikov, D., Weinberger, D., Vijaykumar, D., Rogozińska, D., Herbison, D., Bandy, E., Wang, E., Noland, E., Moreira, E., Senter, E., Eltyshev, E., Visin, F., Rasskin, G., Wei, G., Cameron, G., Martins, G., Hashemi, H., Klimczak-Plucińska, H., Batra, H., Dhand, H., Nardini, I., Mein, J., Zhou, J., Svensson, J., Stanway, J., Chan, J., Zhou, J.~P., Carrasqueira, J., Iljazi, J., Becker, J.,
  Fernandez, J., van Amersfoort, J., Gordon, J., Lipschultz, J., Newlan, J., yeong Ji, J., Mohamed, K., Badola, K., Black, K., Millican, K., McDonell, K., Nguyen, K., Sodhia, K., Greene, K., Sjoesund, L.~L., Usui, L., Sifre, L., Heuermann, L., Lago, L., McNealus, L., Soares, L.~B., Kilpatrick, L., Dixon, L., Martins, L., Reid, M., Singh, M., Iverson, M., Görner, M., Velloso, M., Wirth, M., Davidow, M., Miller, M., Rahtz, M., Watson, M., Risdal, M., Kazemi, M., Moynihan, M., Zhang, M., Kahng, M., Park, M., Rahman, M., Khatwani, M., Dao, N., Bardoliwalla, N., Devanathan, N., Dumai, N., Chauhan, N., Wahltinez, O., Botarda, P., Barnes, P., Barham, P., Michel, P., Jin, P., Georgiev, P., Culliton, P., Kuppala, P., Comanescu, R., Merhej, R., Jana, R., Rokni, R.~A., Agarwal, R., Mullins, R., Saadat, S., Carthy, S.~M., Cogan, S., Perrin, S., Arnold, S. M.~R., Krause, S., Dai, S., Garg, S., Sheth, S., Ronstrom, S., Chan, S., Jordan, T., Yu, T., Eccles, T., Hennigan, T., Kocisky, T., Doshi, T., Jain, V., Yadav, V.,
  Meshram, V., Dharmadhikari, V., Barkley, W., Wei, W., Ye, W., Han, W., Kwon, W., Xu, X., Shen, Z., Gong, Z., Wei, Z., Cotruta, V., Kirk, P., Rao, A., Giang, M., Peran, L., Warkentin, T., Collins, E., Barral, J., Ghahramani, Z., Hadsell, R., Sculley, D., Banks, J., Dragan, A., Petrov, S., Vinyals, O., Dean, J., Hassabis, D., Kavukcuoglu, K., Farabet, C., Buchatskaya, E., Borgeaud, S., Fiedel, N., Joulin, A., Kenealy, K., Dadashi, R., and Andreev, A.
\newblock Gemma 2: Improving open language models at a practical size, 2024.
\newblock URL \url{https://arxiv.org/abs/2408.00118}.

\bibitem[Uppaal et~al.(2023)Uppaal, Hu, and Li]{uppaal2023finetuningneededpretrainedlanguage}
Uppaal, R., Hu, J., and Li, Y.
\newblock Is fine-tuning needed? pre-trained language models are near perfect for out-of-domain detection, 2023.
\newblock URL \url{https://arxiv.org/abs/2305.13282}.

\bibitem[Vapnik(1991)]{NIPS1991_ff4d5fbb}
Vapnik, V.
\newblock Principles of risk minimization for learning theory.
\newblock In Moody, J., Hanson, S., and Lippmann, R. (eds.), \emph{Advances in Neural Information Processing Systems}, volume~4. Morgan-Kaufmann, 1991.
\newblock URL \url{https://proceedings.neurips.cc/paper_files/paper/1991/file/ff4d5fbbafdf976cfdc032e3bde78de5-Paper.pdf}.

\bibitem[Yao et~al.(2023)Yao, Zhao, Yu, Du, Shafran, Narasimhan, and Cao]{yao2023reactsynergizingreasoningacting}
Yao, S., Zhao, J., Yu, D., Du, N., Shafran, I., Narasimhan, K., and Cao, Y.
\newblock React: Synergizing reasoning and acting in language models, 2023.
\newblock URL \url{https://arxiv.org/abs/2210.03629}.

\bibitem[Zhang et~al.(2025)Zhang, Xiao, Liu, Bamler, and Wischik]{zhang2025finetunedlargelanguagemodel}
Zhang, A., Xiao, T.~Z., Liu, W., Bamler, R., and Wischik, D.
\newblock Your finetuned large language model is already a powerful out-of-distribution detector, 2025.
\newblock URL \url{https://arxiv.org/abs/2404.08679}.

\bibitem[Zhou et~al.(2024)Zhou, Xu, Zhu, Zhou, Lo, Sridhar, Cheng, Ou, Bisk, Fried, Alon, and Neubig]{zhou2024webarenarealisticwebenvironment}
Zhou, S., Xu, F.~F., Zhu, H., Zhou, X., Lo, R., Sridhar, A., Cheng, X., Ou, T., Bisk, Y., Fried, D., Alon, U., and Neubig, G.
\newblock Webarena: A realistic web environment for building autonomous agents, 2024.
\newblock URL \url{https://arxiv.org/abs/2307.13854}.

\end{thebibliography}
\bibliographystyle{icml2026}

\newpage
\appendix
\onecolumn
\section{Analysis of Layer-wise Domain Cohesion via Top-K Jaccard Similarity}

This analysis quantifies the evolution of domain-specific representations across the internal layers of a neural network. By processing the four distinct categories of ag\_news dataset —World, Sports, Business, and Sci/Tech—the algorithm identifies how consistently the model activates specific features for samples belonging to the same class. For each layer $l$, the hidden representations are binarized into a set $A$ containing the indices of the $K=10$ most active neurons, such that $A = \{i \mid h_i \in \text{Top-}K(h)\}$. The similarity between any two samples $i$ and $j$ within a batch is then measured using the Jaccard Index:
\begin{equation}
J(A_i, A_j) = \frac{|A_i \cap A_j|}{|A_i \cup A_j|}
\end{equation}

To compute this efficiently for a batch of $N$ samples, the intersection is derived from the product of the binarized feature matrix $B$ and its transpose $BB^\top$, while the union is calculated using the principle of inclusion-exclusion: $|A_i \cup A_j| = |A_i| + |A_j| - |A_i \cap A_j|$. The script isolates the off-diagonal elements of the resulting $N \times N$ similarity matrix to avoid self-comparison bias. For every layer, the algorithm aggregates these pairwise scores into a mean $\mu_l$ and a standard deviation $\sigma_l$. These statistics are visualized as a longitudinal plot across three functional stages: the Lexical Zone (Layers 0–4), the Semantic Trunk (Layers 6–18), and the Specific Zone (Layers 20–25). This visualization serves as a diagnostic tool; a rising mean combined with a tightening standard deviation indicates that the model is successfully distilling diverse inputs into a singular, stable semantic concept as the data moves deeper into the architecture.

\section{Top-$K$ Parameter Sweep and Ablation Analysis}

This experiment evaluated the impact of feature sparsity (controlled by the Top-$K$ parameter) on Out-of-Distribution (OOD) detection across three architectures: Hierarchical Temporal Memory (HTM), First-Order Markov Chains, and RNNs (LSTM). Evaluations were performed using $16k$ Sparse Autoencoder (SAE) embeddings from \texttt{gemma-2-2b}, run on all ag\_news splits with density threshold 0.1.

\subsection{Comparative Performance}
The results (Table \ref{tab:topk_results}) indicate a clear inverse relationship between $K$ and detection accuracy. Peak performance was achieved at the highest sparsity level ($k=10$) for all methods. Notably, the \textbf{First-Order Markov Chain} consistently outperformed or matched more complex architectures, suggesting that local feature transitions are highly discriminative in SAE latent spaces.

\begin{table}[h]
\centering
\caption{Mean AUROC Performance Averaged Across \textbf{ag\_news} all ID Classes}
\label{tab:topk_results}
\begin{tabular}{@{}lccc@{}}
\toprule
\textbf{Top-$K$ Value} & \textbf{HTM (Global)} & \textbf{Markov Chain (1st)} & \textbf{RNN (LSTM)} \\ \midrule
$k=10$  & 0.9238 & \textbf{0.9280} & 0.9131 \\
$k=32$  & 0.9064 & \textbf{0.9195} & 0.9098 \\
$k=64$  & 0.9002 & \textbf{0.9153} & 0.9042 \\
$k=128$ & 0.5000\textsuperscript{*} & \textbf{0.9076} & 0.8788 \\ \bottomrule
\end{tabular}
\vspace{2pt}
\small{\\ \textit{*Note: HTM saturation occurred at $k=128$, resulting in a total loss of discriminative power.}}
\end{table}

\subsection{Key Observations}
\begin{itemize}
    \item \textbf{Optimal Sparsity:} $k=10$ provided the best signal-to-noise ratio. Increasing $K$ likely introduces redundant or "noisy" activations that dilute the class-specific sequence signatures.
    \item \textbf{HTM Saturation:} At $k=128$ (low sparsity), the HTM model reached 1.0 train/val anomaly means immediately. This indicates the synaptic permanence reached a state where every transition was considered "expected," causing the AUROC to collapse to 0.50.
    \item \textbf{Architecture Robustness:} The Markov Chain exhibited the lowest performance decay as $K$ increased, demonstrating superior robustness compared to the RNN and HTM when dealing with denser activation patterns.
\end{itemize}

\section{Impact of Density Filtering on Semantic features}
\begin{figure}[t]
  \centering
  \includegraphics[width=\linewidth]{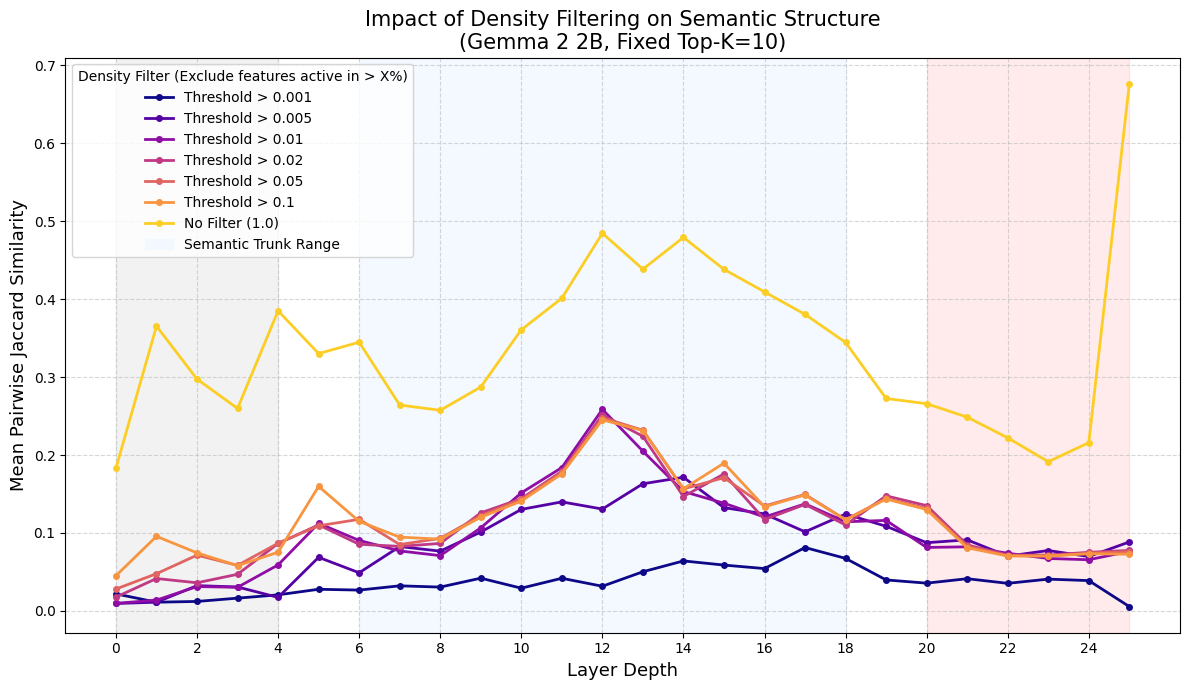}
  \caption{}
  \label{fig:density_sweep}
\end{figure}

To investigate the influence of feature sparsity on internal representations, we performed a parameter sweep over density thresholds while maintaining a fixed top-$K$ activation constraint of $K=10$. Using the Gemma 2 2B model architecture, we computed the Mean Pairwise Jaccard Similarity between binarized sparse distributed representations (SDRs) across all model layers. The density filter serves to exclude ubiquitous features—those active in more than a specified percentage of the PILE dataset—ranging from aggressive filtering ($0.1\%$) to no filtering ($100\%$). Figure \ref{fig:density_sweep}

\end{document}